\documentclass[conference]{IEEEtran}
\IEEEoverridecommandlockouts
\usepackage{cite}
\usepackage{amsmath,amssymb,amsfonts}
\usepackage{algorithmic}
\usepackage{graphicx}
\usepackage{hyperref}
\usepackage{textcomp}
\usepackage{xcolor}
\usepackage{gensymb}
\def\BibTeX{{\rm B\kern-.05em{\sc i\kern-.025em b}\kern-.08em
    T\kern-.1667em\lower.7ex\hbox{E}\kern-.125emX}}

\begin{document}

\title{
Enhancing Robot Learning through Learned Human-Attention Feature Maps
}

\author{
\IEEEauthorblockN{Daniel Scheuchenstuhl*}
\IEEEauthorblockA{\textit{CPS Group} \\
\textit{TU Wien}\\
Wien, Austria \\
0009-0000-6080-2898}
\and
\IEEEauthorblockN{Stefan Ulmer*}
\IEEEauthorblockA{\textit{CPS Group} \\
\textit{TU Wien}\\
Wien, Austria \\
0009-0009-7654-7743}
\and
\IEEEauthorblockN{Felix Resch*}
\IEEEauthorblockA{\textit{CPS Group} \\
\textit{TU Wien}\\
Wien, Austria \\
0009-0004-3240-3725}
\and
\IEEEauthorblockN{Luigi Berducci}
\IEEEauthorblockA{\textit{CPS Group} \\
\textit{TU Wien}\\
Wien, Austria \\
0000-0002-3497-6007}
\and
\IEEEauthorblockN{Radu Grosu}
\IEEEauthorblockA{\textit{CPS Group} \\
\textit{TU Wien}\\
Wien, Austria \\
0000-0001-5715-2142}
\thanks{
*Indicates authors with equal contributions
}

}

\maketitle

\begin{abstract}
Robust and efficient learning remains a challenging problem in robotics, 
in particular with complex visual inputs. 
Inspired by human attention mechanism, with which we quickly process complex visual scenes and react to changes in the environment, 
we think that embedding auxiliary information about focus point
into robot learning would enhance efficiency and robustness of the learning process. 
In this paper, we propose a novel approach to model and emulate the 
human attention with an approximate prediction model. We then 
leverage this output and feed it as a structured auxiliary feature map 
into downstream learning tasks.
We validate this idea by learning a prediction model from
human-gaze recordings of manual driving in the real world.
We test our approach on two learning tasks - 
object detection and imitation learning. 
Our experiments demonstrate that the inclusion of predicted human attention 
leads to improved robustness of the trained models to out-of-distribution samples
and faster learning in low-data regime settings. 
Our work highlights the potential of incorporating structured auxiliary information in representation learning for robotics and opens up new avenues for research in this direction.
All code and data are available online\footnote{Code/Data at \url{https://github.com/CPS-TUWien/learning\_human\_attention}}
\end{abstract}


\section{Introduction}

Robot learning has seen significant progress in recent years, 
developing systems able to perform increasingly complex tasks in a variety of 
challenging environments \cite{andrychowicz2020learning_dext_manipulation, levine2018learning}.
However, the performance of the learning process often depends on the quality of representations,
which retain the important features extracted from high-dimensional sensor data. 
Effective representation learning is therefore crucial for achieving high performance in robot learning tasks,
and an increasing effort has been invested into this fundamental research area
\cite{bengio2013representationlearning, blukis2022spatial_semantic_representation, heravi2022objectawarerepresentations}. 

Most of the modern approaches to representation learning build on self-supervised learning and generative models \cite{pari2021effectiveness_representation_learning, kingma2013vae, goodfellow2020gan},
and have shown promising results
in learning effective representations reusable in many different downstream tasks. 
However, we believe there is still much to be gained by taking inspiration from human behavior. 
Humans are able to process rich visual scenes and perform complex visual-motor tasks with great efficiency, 
due in part to the sophisticated attention mechanisms we employ \cite{johnson2004attention}. 
Attention allows us to capture the most important features to accurately perform a task.

In this work, we build on this insight by developing a new model that mimics the human attention.
However, instead of learning representations 
retaining the most salient features,
we propose to enrich the input data with attentional
feature maps.

To validate this idea, we collect real-world data of human gaze and  
train a model to distill human attention maps from a sequence of visual inputs. 
Having a model able to accurately predict attention maps,
we can use them in unseen frame sequences, 
without any human in the loop.

\vspace{0.5em}

Overall, our work proposes the following contributions:

\begin{itemize}
    \item A novel model to predict human attention maps from visual input, 
    trained on real-world data of human gaze collected during manual driving in scaled miniature cars.   
    
    \item The integration of the learned representations based on human attention 
    into two downstream robot-learning tasks of object detection and imitation learning. 

    \item Experimental evaluation of the effectiveness of our approach by comparing it 
    to existing methods that do not incorporate human-attention features.     
\end{itemize}

In the rest of this paper,
we will explain the proposed methodology
and experimental results.

\begin{figure}[t!]
    \centering
    \includegraphics[width=\columnwidth]{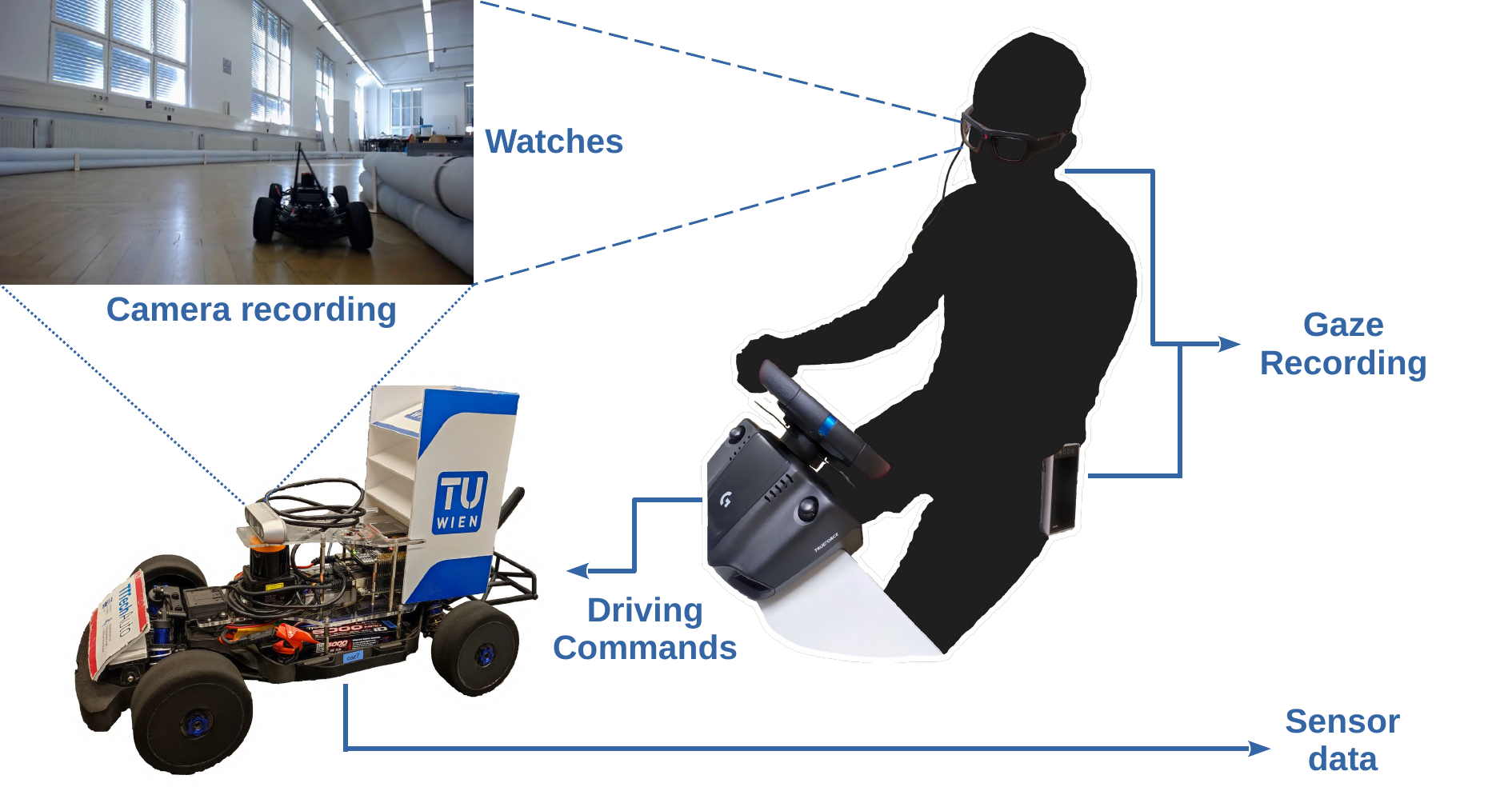}
    \caption{
    Hardware setup to record human-gaze while manually driving a miniature racecar.
    It consists of a eye-tracking system, remote control of the vehicle 
    and data acquisition system.
    }
    \label{fig:hardware_setup}
\end{figure}

\section{Related Works}

In this section, we include the works related to our contributions.
Considering our application in the context of autonomous racing, we include a review of the existing work 
of learning-based approaches used in autonomous racing.

\noindent\textbf{Attention models.}
Research on human attention dates back several decades and has been extensively studied due to its significant impact on learning and perception.
In \cite{johnson2004attention},
the authors discuss the fundamental mechanism of attention and its advantageous effects on learning of humans.
More recently, the notion of focus or attention has been
adopted by the machine learning community \cite{bahdanau2014neural}, 
leading to many successful applications.
Among them, notable mention is the self-attention mechanism
adopted by transformer in natural language processing \cite{vaswani_attention_nodate},
emotion detection \cite{peng_speech_2020},
or image recognition \cite{zhao_exploring_2020}.
However, the connection between artificial attention 
mechanism commonly adopted in modern neural architectures
and the human or biological attention is not clear.
This motivates a new effort in research
to explore how the two are related \cite{lindsay_attention_2020, lindsay_unified_2019}.
In this direction, our contributions
try to demonstrate the practical usability
of human-based features.

\noindent\textbf{Representation Learning.} 
In recent years, there has been significant progress in the field of representation learning \cite{bengio_representation_2013}.
Most of the modern approaches rely on self-supervised
learning and generative models \cite{pari2021effectiveness_representation_learning, kingma2013vae, goodfellow2020gan},
using variants of auto-encoders \cite{wang_auto-encoder_2016, zhang_survey_2022} to learn a low-dimensional
latent representation. 
The applications range from
computer vision \cite{chen_generative_2020, dosovitskiy_image_2021},
natural language \cite{radford_improving_2018},
or multimodal inputs \cite{bao_beit_2022, he_masked_2022}.
Compared with these works,
we do not use attention to provide a compact
representation of the input,
but instead propose to enrich it 
with additional attentional features.

\noindent\textbf{Autonomous Racing.}
Considering our robotics experiments
have been framed in the context of autonomous racing,
we provide a review of existing approaches
with focus on learning applications.
A complete overview of this research field
is provided in \cite{betz_autonomous_2022}.
Despite the wide use of learning-based approaches for perception, 
even in racing competitions \cite{dhall_real-time_2019, strobel_accurate_2020, de_rita_cnn-based_2019},
most of the planning approaches currently deployed on 
hardware cars use either traditional control,
such as model predictive control (MPC), 
or analytical approaches \cite{coulter1992purepursuit, sezer2012followthegap}.
However, an increase effort on learning-based control 
raised in the last years.
Some works use learned models in conjunction with MPC \cite{rosolia2017learning},
in an end-to-end fashion with imitation or reinforcement learning \cite{lee_ensemble_2019, pan_agile_2018, brunnbauer2022latent, cai_vision-based_2021, berducci2021hprs},
or in a hierarchical framework \cite{weiss_deepracing_2020, wadekar_towards_nodate, mahmoud_optimizing_2020} 
where a deep model generates trajectories that are then tracked with a low-level controller.
In \cite{lee_perceptual_nodate}, MPC is combined
with a novel attention mechanism that uses the generated trajectory to identify a region of interest on images.

\section{Human-Attention Model}

We first formalize the problem of reproducing the human-attention mechanism.
Considering input images $x_t \in \mathbb{R}^{m \times n}$,
we define an attention map $y_t \in \mathbb{R}^{m \times n}$
retaining all the focus points obtained by recording the human gaze.

\begin{figure*}[ht]
    \centering
    \includegraphics[width=\textwidth]{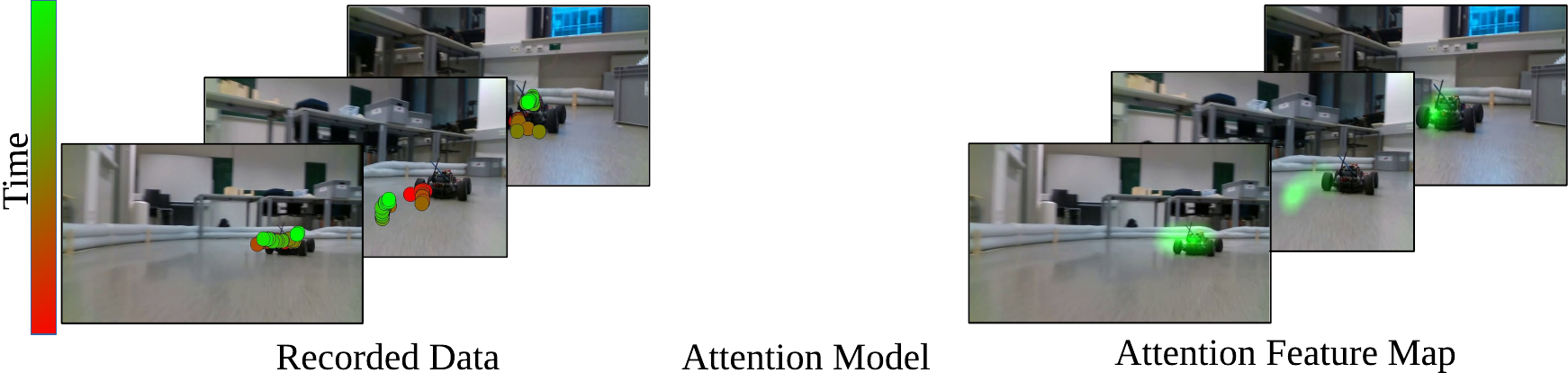}
    \caption{
    The proposed human-attention model is trained to distill attention
    feature maps from visual input and learn the most-relevant focus area.
    }
    \label{fig:attention_model_diagram}
\end{figure*}

The human-attention model $M$ is then defined as
$$
y_t = M(x_t)
$$

Considering the sequential nature of the human-attention mechanism,
which focuses on different areas of the image in a sequential way,
and the limited capabilities of the recording system,
which is able to capture up to a fixed number of points simultaneously,
we propose an approximation of the attention maps with an
exponential-decay time processing.
Let $x_{t} \in \mathbb{R}^{m \times n}$ be the frame captured at time $t$,
$P_t = \{p_{j, t}\}_{j}$ the set of focus points tracked with our system, 
and $h_t \in \mathbb{R}^{m \times n}$ the heatmap computed 
by centering a Gaussian distribution over each focus point $p \in P_t$.
Since $h_t$ only captures the instantaneous attention map of time step $t$,
we aggregate it to produce $y_t$ which resembles the complete attention focus as:
\begin{align*}
y_0 &= h_0 \\
y_t &= max(h_{t}, (1 - r) y_{t-1})
\end{align*}
where $max$ refers to the pixel-wise max operation.
In practice, we use $r=0.17$ to cover all the focus point in the last second
of recording, according to the frequency of our system. We normalize 
the attention maps to have likelihood values between $0$ and $255$.

Having formulated the input and output to model the human-attention
mechanism, we now introduce the learning of an attention predictor
that serves to exploit the use of attention beyond the data-collection setup.
We frame the problem of learning an attention predictor as an image restoration problem which aims to produce approximated attention heatmaps. 
We build on top of the state-of-the-art model U-Net \cite{UNet} and use a smooth L1 loss, as commonly adopted for image segmentation and related application domains.
Specifically, we produce each attention map by centering an isotropic
2D Gaussian distribution with a zero mean and unit standard deviation
over each focus point, and perform the aggregation described above.
The use of Gaussian distribution is a smooth representation which resembles
the likelihood of a driver's attendance to that location. 
The main advantage of it is to have a continuous output space where
adjacent pixels are not independent. 

\textbf{Hardware Setup.}
In this section, we will provide a detailed description of the hardware setup 
used to collect a dataset of human attention while driving a miniature race car on various tracks.
Since the primary objective of this experiment was to analyze human attention and behavior while 
driving, we mounted a camera on the car and asked human drivers to control the car through it by
using either a steering wheel and pedals, or a game controller. 
To ensure that the driver's focus was entirely on the video stream captured by the camera and rely 
solely on it to control the car, 
we seated them away from the track.

To record the driver's gaze, we used a VPS 19 System eye-tracker. 
The eye-tracker recorded the driver's gaze points, which were then transformed into the camera's frame. 
All the gaze points outside the camera frame were discarded, to filter out when the drivers
moved their head or looked around.

The miniature vehicle used in this experiment was an F1Tenth race car
equipped with an inertial measurement unit, a 2D LiDAR, and an RGB camera. 
The LiDAR Hokuyo 10LX could sense distances up to $10$ meters in a $270$-degree field of view at a frequency of $40$ Hz.
To capture a diverse set of images, we used two models of RGB camera, either 
a Logitech C930e or an Intel RealSense d435i. 
Both cameras had a resolution of $848 \times 480$ pixels and a frequency of $30$ FPS.
All the computation to control the car was carried out on a NVIDIA Jetson Xavier NX board. 
The board had six cores and a NVIDIA Volta GPU, making it powerful enough to run the deep models 
for inference later on.
During the experiment, all the gaze data, sensor data, and video streams were recorded and synchronized with the first-person video for further processing.

\textbf{Model training.}
The human-attention model introduced in the previous setting has been trained
on a Tesla T4 GPU with 16 GiB VRAM for 60 epochs using a batch size of 16 and four workers.
We compared various network architectures for semantic segmentation, like
Attention U-Net \cite{AttentionUNet}, U-Net++ \cite{UNetPlusPlus}, Path Aggregation Network (PAN) \cite{PAN} and DeepLabV3+ \cite{DeepLabV3+}. Based on the validation results, U-Net++ with Concurrent Spatial and Channel Excitation (scSE) \cite{scse} performed the best.
During preprocessing, the upper third of the input frames is discarded to bias the learning to the track events.
We use horizontal image flipping and colorspace augmentation, changing brightness, saturation and hue. The human-attention model is not biased towards out-of-distribution illumination perturbations, as the magnitude of change applied for colorspace augmentation is equal to YOLOv7.
The optimization is carried on with AdamW optimizer with a Smooth L1 Loss
and the hyperparameters tuned with PyHopper \cite{PyHopper}. 
We finally select an initial learning rate of $3e^{-4}$ and weight decay of $5e^{-5}$,
and apply automatic mixed precision and learning rate scheduling.

\section{Experiments with the Human-Attention Model}
In the following, we describe the experiments to demonstrate the benefit of integrating human attention in autonomous racing tasks, respectively in object detection and imitation learning.

\noindent\textbf{Object Detection.} 
Considering object detection in the context of F1Tenth autonomous racing,
we consider the detection of the most-common static and dynamic obstacles, 
respectively boxes and other cars.
We collect a training dataset consisting of these object classes,
and compare the performance of the state-of-the-art YOLOv7 \cite{wang2022yolov7} 
trained with the predicted attention (\textbf{Att-YOLOv7})
and without it (\textbf{YOLOv7}).
While YOLOv7 expects a 3-channel input, we modified the Attention-YOLOv7
to additionally receive the attention map, resulting in 4-channels.
We trained the models with SGD using
learning rate $1e^{-2}$, momentum $0.937$, and weight decay $5e^{-4}$. 

\begin{figure}[hb]
    \centering
    \includegraphics[width=\columnwidth]{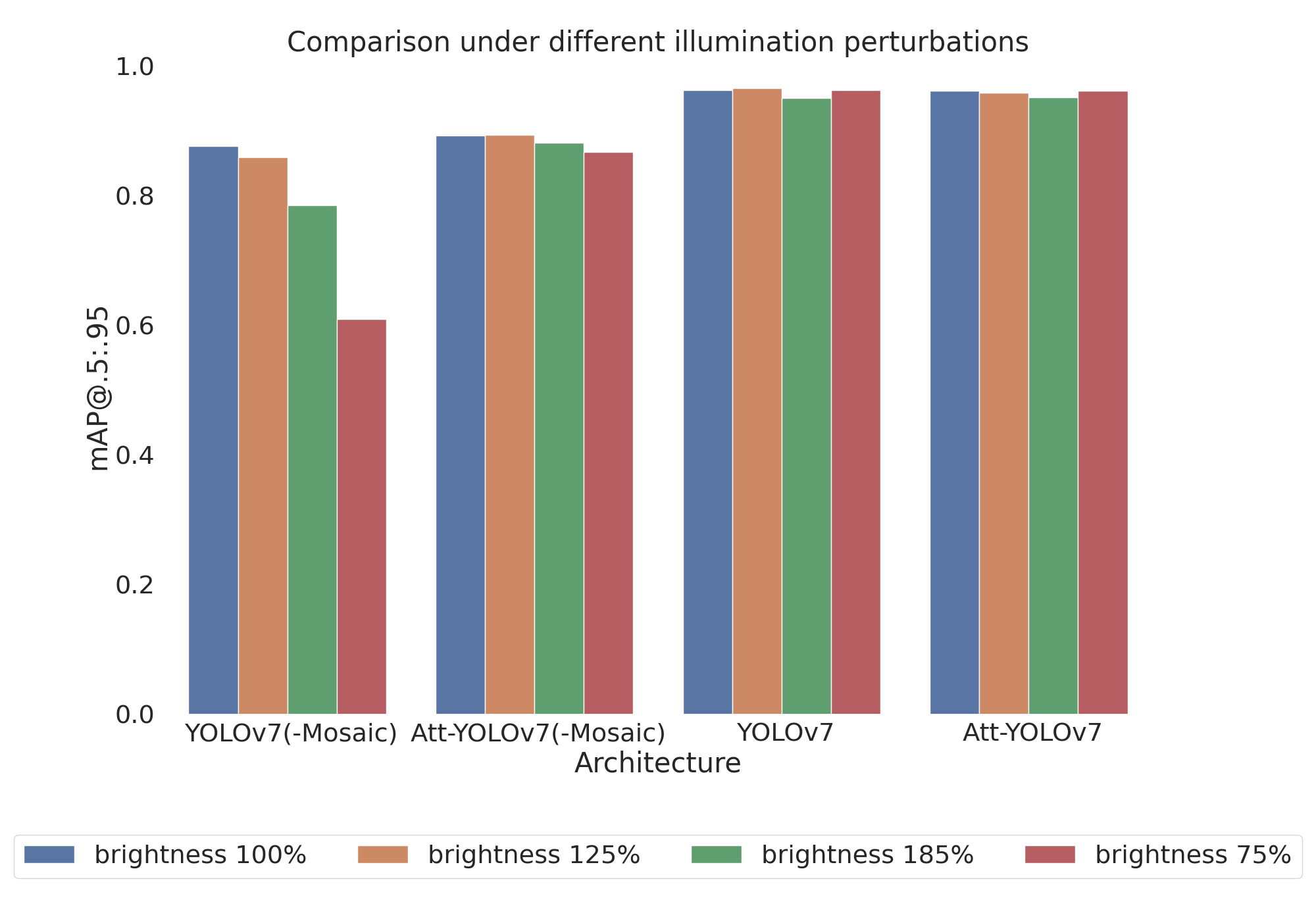}
    \caption{
    Robustness evaluation under different brightness perturbations.
    The horizontal axis shows the various models and ablations,
    and the vertical axis reports the performance in term of mAP@.5:.95 score.}
   \label{fig:object_detection_accuracy_brightness_per_category}
\end{figure}

To analyze the robustness of the trained model to out-of-distribution samples,
we evaluate them on images with heavy-perturbed brightness,
changing it between $75\%$ and $185\%$ of the original value.
We consider brightness perturbations because vision models are sensitive to changes in illumination.
While the data augmentation techniques applied during training
(i.e., horizontal flipping, image translation $\pm 0.2$, scaling $\pm 0.5$, HSV, mosaic) make the model invariant to many transformations,
we still observe high sensitivity to the perturbed inputs.
To assess the impact of various augmentations,
we conduct an ablation study on them and 
report the results for Mosaic in Figure \ref{fig:object_detection_accuracy_brightness_per_category}.
We choose Mosaic because it was the agumentation technique
with largest impact on the model performance.
For each of the two models, 
the performance of training with and without Mosaic (\textbf{$\pm$Mosaic}) is reported as \textit{mAP}, a standard metric for this application.

We observe that the introduction of predicted attention 
as additional input channel produces more robust models
to brightness perturbations.
The performance for models trained with Mosaic are equally good,
but the gap is evident when trained without it.
In particular, YOLO-v7 shows a drop of $10\%$ and $25\%$
for perturbations of $185\%$ and $75\%$ respectively.
Conversely, the model trained with attention exhibits the same level of performance for all the perturbations.
This result suggests that feeding human attention as additional input
facilitates learning robust representations for object detection.





\noindent\textbf{Imitation Learning.} 
We consider the task of imitating the expert driver
in controlling the F1Tenth racecar.
The agent is an end-to-end model
which receives RGB-images and predicts the driving commands
for steering and velocity.

\begin{figure}
    \centering
    \includegraphics[width=\columnwidth]{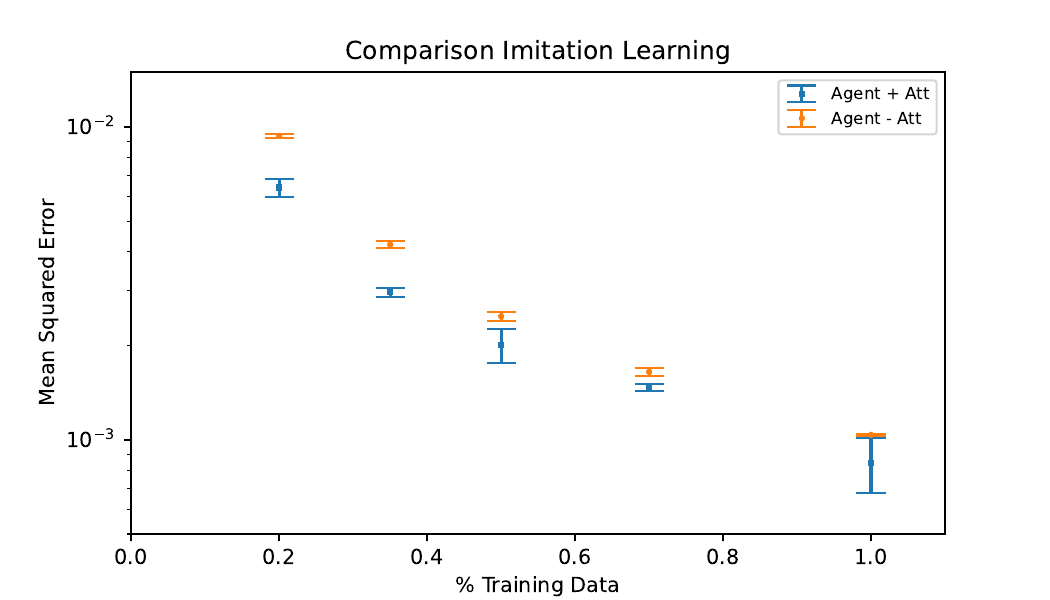}
    \caption{
    Evaluation of prediction error for imitation learning
    under different training budgets 
    ($1.0$=all available $16\,000$ samples). 
    We report the average mean squared error over the last $50$ epochs
    with relative error margins
    for the agent trained with (\textit{Agent+Att}) and without (\textit{Agent-Att}) attention features.
    }
    \label{fig:imitation_learning_error_train_data}
\end{figure}

To experiment with a different integration of human attention
in the input, we mark the attention points in the image,
without adding it as additional channel.
In this experiment, we use the recorded human attention,
and give the marked input to the agent model
which encodes it into an embedding with ResNet18.
The representation is then feed into a series of fully-connected
layers which predict the driving commands.
We compare the agent model trained with and without attention
and report the performance in term of prediction error.
Figure \ref{fig:imitation_learning_error_train_data}
shows the evaluation for different training budgets, 
expressed as a fraction of the available training samples.

We observe a positive impact of marking the input frames
with attention points, especially in low-data regimes.
In fact, while using a large amount of data
makes the performance comparable and the introduction of attention
does not degraded the predictions,
it results impactful 
when the training data are scarce.
This result shows that the use of human attention
has the potential for more-efficient learning.

\section{Conclusions}

In this study, we investigated the impact of human attention on enhancing robot learning in terms of both robustness and efficiency. 
To achieve this goal, we developed a data-collection pipeline to record data from human interaction in a driving task and propose a novel method to approximate human attention using a prediction model.
This information is then used to enrich the visual input
with attentional feature maps.
We assess the impact in two specific learning tasks. 
In the object detection task, 
we observed a significant improvement in 
robustness to out-of-distribution samples when using attention data. 
In the imitation learning task, 
we observed lower prediction error and faster convergence,
especially in low-data regimes.

These promising results highlight the potential of integrating human-based data into machine learning pipelines to improve robot learning. 
We intend to continue investigating this approach in future work to further understand its potential
and explore other possible way to integrate human-based features into learning models. 
Overall, our findings suggest that the integration of human attention data can enhance the robustness and efficiency of machine learning models, which has significant implications for a range of real-world applications.

\section*{Acknowledgment}

Luigi Berducci was supported by the
Doctoral College Resilient Embedded System of TU Wien Informatics.
Felix Resch was supported by the Horizon Europe Research and Innovation programme (Grant Agreement 101070679).

\bibliographystyle{IEEEtran} {
    \bibliography{main}  
}

\vspace{12pt}

\end{document}